\documentclass[12pt,letterpaper]{article}
\usepackage{setspace}
\usepackage[margin=1.25in]{geometry}

\usepackage{amsthm, amsmath, amssymb, xpatch, xcolor, hyperref}
\usepackage[titletoc]{appendix}
\usepackage{enumitem} %
\usepackage{float} 
\usepackage[margin=0.5cm, font={small,it}]{caption}
\usepackage{wrapfig}
\usepackage[normalem]{ulem} 
\usepackage{nicefrac} 

\usepackage{mathtools} 

\usepackage[mathscr]{euscript} 

\usepackage{tocloft} 

\newcommand{\ti}[1]{\textbf{#1}}

\usepackage{mdframed}
\newcommand{\oop}{
	\vspace{5pt}
	\begin{mdframed}[
	linecolor=lightgray, 
	backgroundcolor=gray!4,
	linewidth=1pt,
	innerleftmargin=15pt, innerrightmargin=15pt, innertopmargin=10pt, innerbottommargin=10pt,
]}
\newcommand{\eed}{
	\end{mdframed}
	\vspace{5pt}
}

\newcommand{\ooop}{\begin{itemize}[
	leftmargin=1.5em, 
	labelsep=0.8em
]}
\newcommand{\eeed}{\end{itemize}}

\newcommand{\opp}{\begin{quote}}
\newcommand{\edd}{\end{quote}}

\definecolor{cardinal}{rgb}{0.7, 0.2, 0.2}

\definecolor{sky}{rgb}{0.15, 0.35, 0.75}

\definecolor{bigredc}{rgb}{0.8, 0.0, 0.0}

\definecolor{grayyc}{rgb}{0.6, 0.6, 0.6}

\theoremstyle{plain}
\newtheorem*{theorem*}{Theorem}

\newtheorem*{lemma*}{Lemma}

\theoremstyle{definition}

\newtheorem*{definition*}{Definition}

\newtheorem*{example*}{Example}

\usepackage{graphicx}

\newcommand{\op}{\begin{itemize}
}
\newcommand{\ed}{\end{itemize}}

\newcommand{\ope}{\begin{enumerate}}
\newcommand{\ede}{\end{enumerate}}

\newcommand{\xm}{\item[]}
\newcommand{\im}{\item}

\title{Pessimistic Meta-Induction and Its Limits: \\Lessons from Frequentist Statistics \\and Machine Learning Theory}

\author{Hanti Lin\thanks{{\em AI use statement}: AI tools were used in the preparation of this paper only for English editing, not for generating any content or images.} \\University of California, Davis \\ika@ucdavis.edu}

\date{}

\begin{document}

\maketitle

\begin{abstract} \noindent This paper challenges the pessimistic meta-inductive argument against scientific realism by undermining its inductive step rather than its historical premise. Although related challenges already exist, I develop a new one. Drawing on a general epistemology of scientific inference developed in frequentist statistics, machine learning, and formal epistemology, I evaluate induction in terms of convergence to the truth. I argue that ordinary enumerative induction can achieve everywhere convergence, whereas meta-induction fails even to achieve almost everywhere convergence. Indeed, in the problem context where meta-induction arises, the failure is deeper: no inference method whatsoever achieves almost everywhere convergence.
\end{abstract}


\section{Introduction}

The pessimistic meta-inductive argument against scientific realism is meant to show, roughly, that since all or most past scientific theories turned out to be not even approximately true, our current scientific theory is no exception (Laudan 1981, Putnam 1978).\footnote{Also see Wray (2015) for a reconstruction of different types of the meta-inductive argument.} Many replies challenge the historical premise that most past theories have indeed been shown to be not even approximately true; see, for example, Devitt (1984), Kitcher (1993), Psillos (1994), Worrall (1994), and Leplin (1997). Here, however, I pursue a complementary strategy: undermining the inductive step itself. Relatively few works do so. Notable examples include Lewis (2001) and Magnus and Callender (2004), who appeal to considerations related to the base-rate fallacy to explain why the meta-inductive step does not amount to justified induction.

Without criticizing those allies of mine, I propose a new way to challenge the inductive step---one based on a {\em general} approach to the epistemology of scientific inference, now thriving in many branches of frequentist statistics, especially nonparametric statistics, and central to the theoretical foundations of machine learning. These are, after all, fields devoted to the systematic study of scientific inference---by scientists and for scientists. The core idea is to evaluate inference methods by their properties of {\em convergence to the truth}---whether in ordinary scientific reasoning, statistical procedures, or machine learning algorithms, and whether or not the data are generated stochastically. Although this idea has a long history, traceable to C. S. Peirce (1902), its core theses have only recently been articulated by philosophers with clarity and generality; see, for example, Lin (2025). My aim is to use this framework to explain why, despite their superficial syntactic similarity, ordinary induction and meta-induction differ sharply:
	\op
	\im In the context of ordinary induction---for example, when we ask whether the ravens to be observed in the future will all be black---enumerative induction is justified because it achieves the relatively strong standard of {\em everywhere convergence}, i.e., convergence to the true answer at {\em every} possible state of the world on the table.
	\im In the context of meta-induction---for example, when we ask whether the scientific theories to be tested in the future will all be refuted by data---enumerative induction is not justified because it fails to achieve even the weaker standard of {\em almost everywhere convergence}, i.e., convergence to the true answer at {\em almost all} possible states of the world on the table.
	\ed
The relevant notion of ``almost all'' will be defined rigorously below, though it has already been used in machine learning to evaluate algorithms for causal structure learning (Lin and Zhang 2020). In fact, the case against meta-induction is stronger still. If a problem context is defined by a set of competing hypotheses as potential answers to the question posed, then in the context in which meta-induction is applied, no inference method whatsoever achieves almost everywhere convergence---this is the main mathematical result of the present paper.

I begin with a simple mathematical model of ordinary induction and meta-induction, open to future refinement (Section~\ref{sec-setting}). Though somewhat toy-like, it captures an appealing version of meta-induction and is therefore worth undermining on its own terms. Its simplicity also helps reveal the underlying structure of the problem and suggests how the result might be extended. Along the way, I explain the epistemological ideas in use, especially the concept of convergence to the truth, while avoiding some of its more traditional and misleading formulations (Section~\ref{sec-formal}). I conclude by emphasizing that this epistemology is a very general framework for scientific inference---one developed largely by scientists, with help from a few philosophers, and broad enough to encompass much of frequentist statistics and machine learning theory (Section~\ref{sec-closing}).

\section{Setting}\label{sec-setting}

Consider the data sequence: 
$$
\mathtt{001}\;\mathtt{000001}\;\mathtt{000000000} \,,
$$
where each $\mathtt{0}$ means nothing bad---our current scientific theory passes a new test---and each $\mathtt{1}$ means something bad---our current theory fails a test and must be replaced. The first block, $\mathtt{001}$, shows that our first theory failed quickly, after just three tests. We then moved to a second theory, which also failed but survived longer, as represented by the longer block $\mathtt{000001}$. Our third theory, by contrast, is still doing well: the third block has not (yet) ended with $\mathtt{1}$.

More generally, let $T_n$ be the $n$-th theory we would develop to maturity if we were diligent enough (and lived long enough). The $n$-th occurrence of $\mathtt{1}$ in the data sequence marks the downfall of $T_n$ and its replacement by $T_{n+1}$.

Now imagine that the following data sequence represents our current state of evidence:
	\oop   
	$
	\mathtt{000001}\;\mathtt{00000001}\cdots\cdots \mathtt{0000000001}\;\overbrace{\mathtt{00000000000000000000000000000000000}}^{\text{$T_{n+1}$ has passed many tests.}}\,
	$
	
	\quad\! $\uparrow$ \quad\quad\quad\,\, $\uparrow$ \quad\quad\quad\quad\quad\quad\quad\,\! $\uparrow$
	
	\quad $T_1$ fails. \; $T_2$ fails. \quad\quad\quad\quad\,\! $T_{n}$ fails, where $n$ is large.
	\eed  
Confronted with this data sequence, we might be tempted to draw one of two types of inductive inference. The first is:
\oop  
	\ti{Ordinary Induction.} Consider the annotations {\em above} the data sequence: the current theory $T_{n+1}$ has undergone many tests and survived them all. We infer---inductively and ordinarily---that $T_{n+1}$ would never suffer a downfall, no matter how many further tests it faces.
\eed   
The second is suggested by Putnam (1978) and Laudan (1981):
\oop 
	\ti{Pessimistic Meta-Induction.} Consider the annotations {\em below} the data sequence. The first $n$ theories, $T_1, \ldots, T_{n}$, have failed, and $n$ is very large. We infer---inductively and pessimistically---that the current theory $T_{n+1}$, along with all its successors, would also fail if tested sufficiently.
\eed 

Now, which type of induction should we adopt? This is not easy to answer, since both instantiate the same syntactic template:
\oop   
\ti{Syntactic Template of Induction}
\\[0.5em]\emph{Premise 1}. Many \emph{F}s have been observed.
\\[0.3em]\emph{Premise 2}. All of them are \emph{G}s.
\\------------------------------------------------------
\\[-0.2em]\emph{Conclusion}. So, all \emph{F}s are \emph{G}s. 
\eed   

To break the symmetry between ordinary induction and pessimistic meta-induction, we must look beyond their syntactic form. If I am right, there is an important asymmetry:
\oop 
\ti{The Disparity Thesis} ({\em To Be Refined and Defended Below}).
\ooop
\im In the context where the question is whether the current theory would fail if sufficiently tested, the ordinary inductive method makes evidence an indicator of the true answer. 
\im But in the context where the question is whether every scientific theory in the pipeline would fail if sufficiently tested, {\em every} non-deductive method---including the pessimistic meta-inductive method---{\em fails} to make evidence an indicator of the true answer.  
\eeed
\eed 

Certain concepts must be defined rigorously before this disparity thesis can be refined into a theorem. I will appeal to the concept of convergence and to a topological conception of ``almost everywhere''.

But even setting aside the technical details, the disparity should not be surprising at an intuitive level. Let infinite binary data sequences represent possible states of the world. These states correspond, more or less, to real numbers in the unit interval, expressed in their binary expansions. Let us ignore the minor redundancy of binary representation (for example, both $\mathtt{1\bar{0}}$ and $\mathtt{0\bar{1}}$ represent the same real number, namely $0.5$). In one problem context, the question is whether the current theory would fail. Equivalently, we ask whether the actual state of the world is a particular real number $r$ in the unit interval, where $r$ is the available data sequence concatenated with an infinite block of $\mathtt{0}$s. In a different context, the question is whether every scientific theory in the pipeline would fail if sufficiently tested. The true answer is ``yes'' iff there are infinitely many $1$s in the actual state of the world, that is, iff the actual state is an irrational number in the unit interval.

So the two problem contexts are very different. One asks whether the actual real number is a specific one; the other asks whether it is irrational. As more digits of data arrive, we get more precise information about where the actual state lies in the continuum of the unit interval---an ever-shrinking nonempty interval $[a_n, b_n]$ of real numbers. Thus, in the first case, more data can still refute at least one of the two candidate answers. In the second case, however, no possible data can refute either of the two candidate answers---the two potential answers, ``Rational Number'' vs. ``Irrational Number'', are too closely intertwined on the real line. That is, any nonempty interval overlaps both. So, the latter question is intuitively harder. The formal development below vindicates this intuition.


\section{Formal Development}\label{sec-formal}

It is time to turn the disparity thesis stated above into a theorem.

\subsection{Epistemic Scenarios and Convergence}

An {\em inference method} is a mathematical function that takes a finite data sequence as evidential input and outputs a proposition as a conclusion, subject to revision as more data arrive. A non-deductive method is justified only if it can, in some sense, serve as a good indicator of truth---and the challenge is to make this idea precise. An initial thought is that a good indicator of truth should point to a {\em truth} across a certain {\em range} of {\em epistemic scenarios}. But what truth? What scenarios? And what range? I will explain in turn.

Let us begin with truth. In any problem context, a question is posed, whose potential answers serve as the competing hypotheses. The truth pursued is the (unknown) true answer to that question in context.

Now turn to scenarios. An {\em epistemic scenario} can be modeled by an ordered pair $(s, n)$. The first component $s$ is a {\em s}tate of the world---a coarse-grained possibility that determines the truth value of each competing hypothesis in context. The second component $n$ is a \emph{n}umber (a positive integer) representing an amount of evidence. Accordingly, $(s, n)$ is the epistemic scenario one would be in if the actual state of the world were $s$ and the amount of available evidence were $n$.

Then imagine a particular scenario $(s, n)$ as a point in a two-dimensional space, where $s$ is the $X$-coordinate and $n$ the $Y$-coordinate. The $X$-axis consists of possible states of the world---specifically those compatible with the {\em background assumptions} of the problem context. The $Y$-axis consists of positive integers.

Different problem contexts may require different mathematical modelings of states of the world. For the inductive problems at hand, a particularly simple definition suffices. Let a state of the world $s$ be an infinite binary sequence. Here is an example:
	$$
	s^* \;=\; \mathtt{010101} \cdots \text{(repeating the pattern of \texttt{01})} \cdots 
	$$
The state is formalized as an infinite sequence, but this is not meant to represent a state in which we are immortal and endlessly accumulate data. Rather, the infinite sequence is a convenient device for encoding counterfactuals: if the amount of evidence {\em were} $n$, the evidence {\em would} be the initial segment of that sequence of length $n$. In this article, the states of the world {\em on the table} are all the infinite binary sequences---we make no background assumptions that rule out any sequence. (In other contexts of inquiry, we might have strong background assumptions, which leave few states of the world on the table.) 

Consider, for example, this epistemic scenario: $(s^*, 4)$, where $s^* = \mathtt{010101}\cdots$; here the available evidence is $\mathtt{0101}$ (four data points). Here $T_2$ has just failed its second test, and worse, every theory would fail if sufficiently tested---specifically, if tested at least twice. If the question posed is whether every theory would fail if sufficiently tested, then the true answer in state $s^*$ is $\mathtt{Yes}$, and thus an inference method $M$ outputs the truth at this scenario $(s^*, 4)$ iff $M(\mathtt{0101}) = \mathtt{Yes}$.

Now, what could count as a good indicator of truth? This is not easy to state precisely. But a guiding intuition is that a non-deductive method counts as a good indicator of truth in a problem context only if it outputs the true answer at each of a ``wide range'' of epistemic scenarios. At a minimum, such a range must include epistemic scenarios $(s, n)$ with very large $n$, i.e., with very large amounts of evidence. The idea is that a good indicator of truth must output the truth at least in evidentially favorable scenarios $(s, n)$---with very large $n$---and, ideally, at many states $s$ on the table, if not all. 

One tentative way to picture this idea---before formalizing it---is as follows: a non-deductive method $M$ counts as good only if we can draw a bar across the two-dimensional plane of epistemic scenarios, cutting through every vertical line and dividing the plane into upper and lower regions, such that $M$ outputs the truth across all scenarios in the upper region. This can be formalized as follows:
\oop 
	\ti{Definition (Everywhere Convergence to the Truth).} An inference method $M$ is said to achieve the standard of {\em everywhere convergence to the truth} in a problem context $c$ iff, in that context $c$,
	\ooop 
	\xm for each state $s$ of the world \uline{on the table} (i.e. on the $X$-axis),\\ 
	$\phantom{\;}$ there exists a finite amount of evidence $N$ (on the $Y$-axis) such that, \\
	$\phantom{\;\;\;}$ $M$ outputs the true \uline{answer} at epistemic scenario $(s, n)$ for each $n \ge N$.
	\eeed 
\eed 
The two underlines serve as a reminder of context-sensitivity: what counts as a state of the world on the table is one compatible with the background assumptions in the problem context $c$ at hand, and what counts as a potential answer depends on the question posed in the problem context $c$ at hand. 

{\em Housekeeping.} The evaluative standard just defined is not particularly high; it is often called {\em pointwise convergence}. A higher standard is obtained by swapping the first two quantifiers---`for each $s$' and `there exists an $N$'---yielding {\em uniform convergence to the truth}. Intermediate standards can also be defined. It is natural to require that a non-deductive inference method be justified only if it achieves the highest achievable standard, and I will adopt this requirement later. For now, however, I focus on a minimum qualification for an indicator of truth, so the bar is not raised too high. Indeed, I think the standard just defined---everywhere convergence---is still too strong to serve as a minimum qualification. To define a genuine minimum, the quantifier `for each state' must be weakened to `for almost all states' in a rigorous sense. Nevertheless, I begin with everywhere convergence to sketch the big picture before refining it.

It is important to keep in mind that we are not assessing inference methods as in decision theory. We are not choosing an inference method as a course of action based on its possible outcomes in the remote future, beyond our lifespans. Such a decision-theoretic approach would be pointless, since in the long run we are all dead.

What we do here is instead a form of {\em modal epistemology}. We seek a minimum qualification---a necessary condition---for a non-deductive method to be justified, by evaluating its truth-seeking performance {\em across a range of epistemic scenarios}, including those in which the evidence is extremely favorable. Anything that counts as an indicator of truth should, if possible, point to the truth {\em at least} in such favorable cases. Accordingly, everywhere convergence, or variants of it, will be used to state a necessary condition for a non-deductive method to be justified.

Let me reiterate: here we are doing modal epistemology, not decision theory. The use of convergence in epistemology is not new. Reichenbach (1938) called it a {\em pragmatic} approach, and Schulte (1999) a {\em means-ends} epistemology. Both labels are misleading, as they suggest a decision-theoretic approach. If this were decision theory, it would be a bad one. Instead, it is modal epistemology: whether an inference method is justified depends on its truth-seeking performance {\em across a certain range of possible scenarios}.

\subsection{A Sketch of the Main Result}\label{sec-explain}

The ordinary inductive problem and the meta-inductive problem are two very different contexts. The highest achievable standards in these contexts can be illustrated by the following rough diagram, though more precise statements will be developed as we proceed:
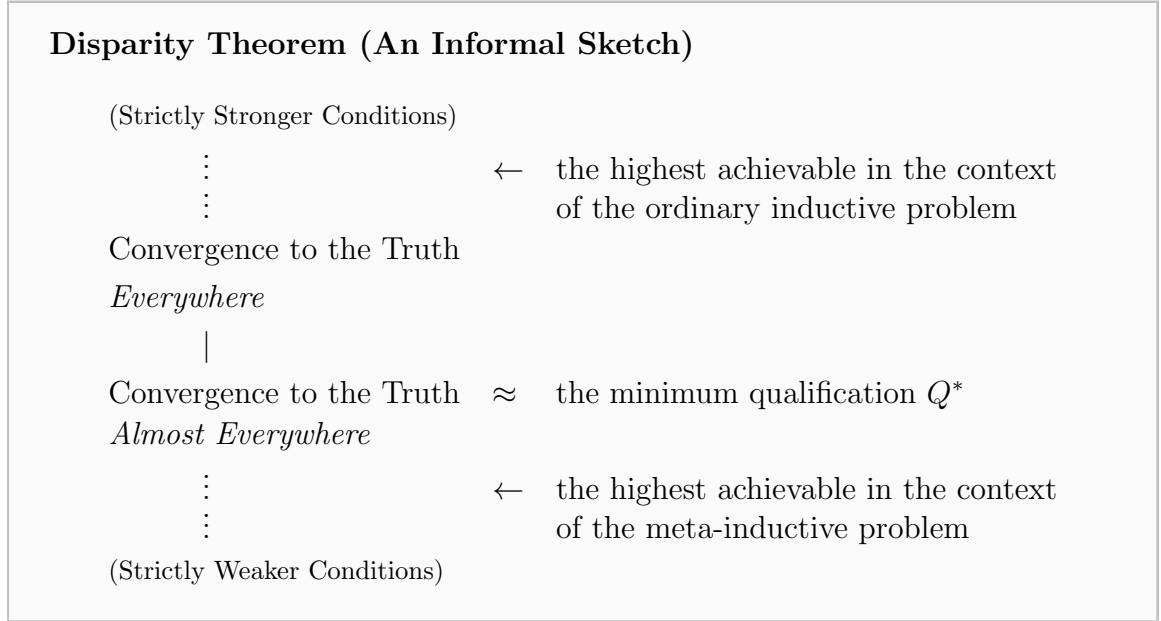
\oop 
\ti{Disparity Theorem (An Informal Sketch)}
\begin{center}
\begin{tabular}{lll}
\footnotesize{(Strictly Stronger Conditions)}
\\
$\qquad\quad \vdots$ & $\leftarrow$ & the highest achievable in the context
\\[-0.5em]
$\qquad\quad \vdots$ &&  of the ordinary inductive problem
\\[-0.2em]
Convergence to the Truth
\\
{\em Everywhere}
\\
$\qquad\quad |$
\\
Convergence to the Truth & $\approx$ & the minimum qualification $Q^*$ 
\\[-0.2em]
{\em Almost Everywhere} &&
\\
$\qquad\quad \vdots$ & $\leftarrow$ & the highest achievable in the context
\\[-0.5em]
$\qquad\quad \vdots$ && of the meta-inductive problem 
\\[-0.2em]
\footnotesize{(Strictly Weaker Conditions)} 
\end{tabular}
\end{center}
\eed 
Again, an intuitively appealing conception of ``almost everywhere'' will be defined rigorously below. What matters for now is that this diagram displays a hierarchy of standards for assessing inference methods, using various modes of convergence as high or low standards for a good indicator of truth. It also indicates where a problem context $c$ ``lies'' along the hierarchy. This is marked by arrows `$\leftarrow$'. Each arrow corresponds to a mathematical theorem indicating the highest standard achievable in a given context. Note the {\em vertical difference} between the two arrows---between the highest achievable standards in the two contexts. This difference is as stark as night and day if the minimum qualification $Q^*$ for a justified non-deductive inference method falls between them.

This choice of $Q^*$---the minimum qualification for justified inference---is not {\em ad hoc}. For nonparametric regression and model selection in statistics, and for supervised learning in machine learning, many algorithms are justified by showing that they achieve (a stochastic version of) everywhere convergence to the truth, where the truth is the true answer to the question posed in context. More on this in the final section.

So, here is my reply to Laudan. It is mistaken to assume that if ordinary induction is justified, then meta-induction must be as well, merely because of their syntactic similarity. Context matters: ordinary induction is justified in its own context, while meta-induction is {\em not} justified in its own context. More specifically, {\em even if} Laudan is right that most empirically successful past theories have later been shown to fail (being not even approximately true), and {\em even if} we strengthen this premise of Laudan's meta-induction by replacing ``most'' with ``all'', his non-deductive inference is still not justified in its context---because every non-deductive inference is doomed to achieve only a very low standard and is thus unjustified in that context.

\subsection{Defining ``Almost Everywhere''}\label{sec-top}

It is time to define ``almost everywhere'' rigorously. 

Think about possible states of the world as points in a space. Consider the following set, where $s_n$ is a state identical to the actual one except that the Planck constant in the Schr\"odinger equation differs from the actual value by $\nicefrac{1}{n}$ (in units of your choice):
$$S = \big\{s_n: n \text{ is a positive integer}\big\} \,.$$
Arguably, this set $S$ of states comes arbitrarily close to the actual state of the world.

Let us return to our setting of possible states as infinite binary sequences. What would be a natural relation of arbitrary closeness for such sequences? Consider the set consisting of the following binary sequences:
	\begin{center}
	\begin{tabular}{c}
	\texttt{011111} $\cdots$ \\
	\texttt{001111} $\cdots$ \\ \texttt{000111} $\cdots$ \\ \texttt{000011} $\cdots$ \\
	$\vdots\qquad$
	\end{tabular}
	\end{center}
This set seems to contain enough initial segments $\mathtt{0}, \mathtt{00}, \mathtt{000}, \ldots$ to approximate the following binary sequence as closely as we wish:
	\begin{center}
	\begin{tabular}{c}
	\texttt{000000} $\cdots$ 
	\end{tabular}
	\end{center}
More generally:
\oop 
\ti{Definition (Cantor Space Topology).} On any set $X$ of infinite binary sequences, the {\em Cantor space topology} provides a natural relation of arbitrary closeness, defined as follows: a subset $S \subseteq X$ gets arbitrarily close to a point $s = e_1 e_2 \cdots \in X$ iff every initial segment $e_1 \cdots e_n$ of $s$ appears as the initial segment of some sequence in $S$.
\eed 
Here, the primitive topological concept in use is not that of open sets, but an equivalent one---that of arbitrary closeness (Arkhangel'skii and Fedorchuk 1990). The Cantor space topology will be the one in use for the rest of this paper.  

Using arbitrary closeness as the core concept, it is very easy to define a geometric conception of ``almost everywhere'' (much easier than the standard textbook treatment):
\oop 
	\ti{Definition (Almost Everywhere, Almost All).} 
	Let $X$ be a set of states of the world. A set is said to cover {\em almost everywhere} in $X$ iff it is big enough to include a subset $S$ of $X$ that stands in the following asymmetric relations to its complement $\overline{S}$ within $X$:  
	\op 
	\im[1.] $S$ comes arbitrarily close to every point in $\overline{S}$;
	\im[2.] $\overline{S}$ comes arbitrarily close to no point in $S$. 
	\ed 
	A property is said to apply to {\em almost all} elements of $X$ iff it applies to every element of a set that covers almost everywhere in $X$. 
\eed 
This definition should look intuitively appealing. Given clauses~1 and~2, the set $S$ does seem to cover almost everywhere. (In fact, clause~1 says that $S$ is dense, and clause~2 says that $S$ is open.) If $S$ covers almost everywhere, then so do its supersets. One more definition:\footnote
	{See Belot (2013) and Lin (2022) for existing epistemological applications of almost-everywhere convergence and related concepts.}

\oop
\ti{Definition (Almost Everywhere Convergence to the Truth).} Within a problem context $c$, an inference method $M$ is said to converge to the truth almost everywhere iff, for each competing hypothesis $h$ as a potential answer to the question posed in context $c$, $M$ converges to the truth at almost all $h$-states, where
	\ooop 
	\im an {\em $h$-state} is a state on the table at which $h$ is true,
	\im $M$'s {\em convergence to the truth} at a state $s$ means that there exists a positive integer $N$ such that $M$ outputs the true answer at epistemic scenario $(s, n)$ for each $n \ge N$.
	\eeed 
\eed 

Then we have the main result:
\oop 
\ti{Disparity Theorem (Official Version).} In the context of the ordinary inductive problem, there exists an inference method that achieves the standard of everywhere convergence to the truth. In the context of the meta-inductive problem, there exists no inference method that achieves even the lower standard of almost everywhere convergence to the truth.
\eed 
This refines the disparity thesis stated above in response to the pessimistic argument. See the appendix for a proof. 

\section{Closing: Toward a General Account of Scientific Inference}\label{sec-closing}

This paper assumes an epistemological view called {\em achievabilism}:
\oop 
\ti{Two Principles of Achievabilism}
\op[leftmargin=1.7em] 
\im[(1)] There exist justified non-deductive methods in a problem context $c$ if, and only if, the minimum qualification $Q^*$ for a non-deductive indicator of truth is achievable in that context $c$---pending a specification of the minimum qualification $Q^*$.

\im[(2)] A non-deductive method $M$ is justified in a problem context $c$ only if $M$ achieves not merely $Q^*$ but also the highest standard achievable in that context $c$ for a good non-deductive indicator of truth (provided that such a highest achievable standard exists uniquely in $c$)---pending a specification of the correct hierarchy of standards.
\ed
\eed 
These two principles provide only a framework, with two moving parts marked by the `pending' clauses. They leave open both the minimum qualification and the correct hierarchy of standards.

Now, if the minimum qualification $Q^*$ for justified non-deductive inference is everywhere or almost everywhere convergence to the truth, then the Disparity Theorem of this paper suggests the following. In using meta-induction to argue against scientific realism, Laudan aimed to place us in a skeptical context where scientific anti-realism appears plausible. But he unintentionally created a far more skeptical context, one in which all non-deductive methods are unjustified---including his meta-induction.

The first achievabilist principle~(1) is, as far as I know, new. The second achievabilist principle~(2) is not: variants were recently stated by Lin (2022, 2025), though those formulations mistakenly omit reference to the minimum qualification $Q^*$. Lin (2025) traces the idea of achievabilism to Putnam (1965), but an earlier version already appears in Neyman and Pearson (1936). They distinguish two evaluative standards for hypothesis-testing procedures and argue that, in the so-called two-sided problem, only the lower standard should be applied because the higher one is unachievable, though the higher standard should still be applied when achievable.

Convergence as a source of evaluative standards is employed across many scientific fields concerned with scientific inference. Begin with frequentist point estimation, which asks: what is the value of a quantity of interest? A stochastic version of everywhere convergence to the true answer---{\em estimation consistency}---has long been regarded, under Fisher's influence, as a basic requirement for a good estimator (Fisher 1925). See Lehmann and Casella (2006) for a textbook presentation.

In frequentist nonparametric regression, the question becomes: which curve $y = f(x)$ on the $XY$-plane, within a function class $\mathscr{C}$, has the highest predictive accuracy? Here too, a stochastic version of everywhere convergence to the true answer---{\em regression consistency}---is treated as a basic requirement of a good method. This extends point estimation naturally: the target is now a point in a much larger space, namely a curve in a large function class. See Gy\"orfi et al. (2002) for a textbook presentation.

Now modify the $XY$-plane: let $X$ be a class of images and $Y$ the set of two categories ``Yes, it is an image of a cat'' and ``No, it is not.'' Then nonparametric regression becomes classification in machine learning. The question remains structurally the same: which ``curve'' $y = f(x)$ on this ``$XY$-plane'', within a function class $\mathscr{C}$, has the highest predictive accuracy? Again, a stochastic version of everywhere convergence to the true answer is often taken as a basic requirement of a good learning algorithm. This is a pointwise mode of convergence; the stronger uniform mode corresponds to the well-known criterion of PAC learning. See Shalev-Shwartz and Ben-David (2014) for a textbook presentation.

In the non-stochastic setting adopted in this paper, everywhere convergence to the truth has long been studied in machine learning theory, beginning with Putnam (1965) and Gold (1967). See Case and Jain (2017) for a survey.

In another area of machine learning, causal discovery, the question is: which causal model on the table is true, assuming one of them is? Here too, modes of convergence to the truth are among the standard criteria for evaluating learning methods (Spirtes et al. 2000). Everywhere convergence to the truth is called {\em model-selection consistency}, and almost everywhere convergence has also been studied and applied in recent years (Lin and Zhang 2020).

So my appeal to convergence in replying to the pessimistic meta-inductive argument is not a parochial use of an arcane mathematical tool. Instead, it seeks to learn from an epistemological tradition that traces back to Peirce (1902), has taken firm root in the sciences, and is now emerging as a general account of scientific inference.

\section*{Appendix: Proof of the Disparity Theorem}

The first part, concerning the ordinary inductive problem, is elementary in formal learning theory; a pictorial proof is given in Lin (2025).

For the second part, the key ingredients are the Baire Category Theorem and Belot's (2013) main theorem, which he states for a problem context isomorphic to the meta-inductive one. Belot's result implies that any open-minded inference method converges to the truth only on a meager set. Although Belot formulates the theorem for Bayesian methods, the proof extends straightforwardly to the broader class of inference methods considered here, including those with qualitative outputs. 

Now suppose, for {\em reductio}, that almost-everywhere convergence to the truth is achievable, by some inference method $M$. By the first clause in the definition of ``almost everywhere'', $M$ must converge to the truth on a dense subset of the set of $h$-states, for each hypothesis $h$. But denseness for both hypotheses immediately yields Belot's open-mindedness condition. So, by Belot's theorem, $M$ converges to the truth only on a meager domain. This contradicts the Baire Category Theorem, since in Cantor space no meager set can cover almost everywhere. Therefore, almost-everywhere convergence is not achievable. Q.E.D.

\section*{References}

\begin{description}
\im Arkhangel'skii, A. V., \& Fedorchuk, V. V. (1990). The basic concepts and constructions of general topology. In A. V. Arkhangel'skii \& L. S. Pontryagin (Eds.), \textit{General topology I: Basic concepts and constructions. Dimension theory} (pp. 1--55). Springer-Verlag.

\im Belot, G. (2013). Bayesian orgulity. \textit{Philosophy of Science, 80}(4), 483--503.

\im Case, J., \& Jain, S. (2017). Connections between inductive inference and machine learning. In C. Sammut \& G. I. Webb (Eds.), \textit{Encyclopedia of machine learning and data mining} (pp. 261--272). Springer.

\im Devitt, M. (1984). \textit{Realism and truth}. Princeton University Press.

\im Fisher, R. A. (1925). \textit{Statistical methods for research workers}. Oliver \& Boyd.

\im Gold, E. M. (1967). Language identification in the limit. \textit{Information and Control, 10}(5), 447--474.

\im Gy\"orfi, L., Kohler, M., Krzy\.zak, A., \& Walk, H. (2002). \textit{A distribution-free theory of nonparametric regression}. Springer.

\im Kitcher, P. (1993). \textit{The advancement of science}. Oxford University Press.

\im Laudan, L. (1981). A confutation of convergent realism. \textit{Philosophy of Science, 48}(1), 19--49.

\im Lehmann, E. L., \& Casella, G. (2006). \textit{Theory of point estimation}. Springer Science \& Business Media.

\im Leplin, J. (1997). \textit{A novel defense of scientific realism}. Oxford University Press.

\im Lewis, P. J. (2001). Why the pessimistic induction is a fallacy. \textit{Synthese, 129}(3), 371--380.

\im Lin, H. (2022). Modes of convergence to the truth: Steps toward a better epistemology of induction. \textit{The Review of Symbolic Logic, 15}(2), 277--310.

\im Lin, H. (2025). Convergence to the truth. In K. Sylvan, E. Sosa, J. Dancy, \& M. Steup (Eds.), \textit{The Blackwell companion to epistemology} (3rd ed.). Wiley Blackwell.

\im Lin, H., \& Zhang, J. (2020). On learning causal structures from non-experimental data without any faithfulness assumption. \textit{Proceedings of Machine Learning Research, 117}, 554--582.

\im Lyons, T. D. (2002). Scientific realism and the pessimistic meta-modus tollens. In S. Clarke \& T. D. Lyons (Eds.), \textit{Recent themes in the philosophy of science: Scientific realism and commonsense} (pp. 63--90). Springer.

\im Magnus, P. D., \& Callender, C. (2004). Realist ennui and the base rate fallacy. \textit{Philosophy of Science, 71}(3), 320--338.

\im Neyman, J., \& Pearson, E. S. (1936). Contributions to the theory of testing statistical hypotheses: Part I. \textit{Statistical Research Memoirs, 1}, 1--37.

\im Peirce, C. S. (1902). Validity. In J. M. Baldwin (Ed.), \textit{Dictionary of philosophy and psychology}. Macmillan. 

\im Psillos, S. (1994). A philosophical study of the transition from the caloric theory of heat to thermodynamics: Resisting the pessimistic meta-induction. \textit{Studies in History and Philosophy of Science, 25}(2), 159--190.

\im Putnam, H. (1965). Trial and error predicates and the solution to a problem of Mostowski. \textit{Journal of Symbolic Logic, 30}(1), 49--57.

\im Putnam, H. (1978). \textit{Meaning and the moral sciences}. Routledge.

\im Reichenbach, H. (1938). \textit{Experience and prediction: An analysis of the foundation and the structure of knowledge}. University of Chicago Press.

\im Schulte, O. (1999). Means-ends epistemology. \textit{The British Journal for the Philosophy of Science, 50}(1), 1--31.

\im Shalev-Shwartz, S., \& Ben-David, S. (2014). \textit{Understanding machine learning: From theory to algorithms}. Cambridge University Press.

\im Spirtes, P., Glymour, C., \& Scheines, R. (2000). \textit{Causation, prediction, and search} (2nd ed.). MIT Press.

\im Worrall, J. (1994). How to remain (reasonably) optimistic: Scientific realism and the ``luminiferous ether''. \textit{PSA: Proceedings of the Biennial Meeting of the Philosophy of Science Association, 1994}(1), 334--342.

\im Wray, K. B. (2015). Pessimistic inductions: Four varieties. \textit{International Studies in the Philosophy of Science, 29}(1), 61--73.
\end{description}

\end{document}